\RequirePackage{fix-cm}
\documentclass[smallextended]{svjour3}       
\smartqed  
\usepackage{amsmath,amsfonts,amssymb}
\usepackage{graphicx}
\usepackage{setspace}
\usepackage{tocloft}
\usepackage{subcaption}
\usepackage{mathtools}
\usepackage{booktabs}
\begin{document}

\title{Superpixels Based Segmentation and SVM Based Classification Method to Distinguish Five Diseases from Normal Regions in Wireless Capsule Endoscopy}


\author{Omid Haji Maghsoudi}


\institute{Omid Haji Maghsoudi\at
              Department of Bioengineering, College of Engineering, Temple University, Philadelphia, PA, USA, 19122  \\
              Tel.: +1-267-432-6386\\
              \email{o.maghsoudi@temple.edu}
}

\date{Received: date / Accepted: date}

\maketitle

\begin{abstract}
Wireless Capsule Endoscopy (WCE) is relatively a new technology to examine the entire GI trace. During an examination, it captures more than 55,000 frames. Reviewing all these images is time-consuming and prone to human error. It has been a challenge to develop intelligent methods assisting physicians to review the frames. The WCE frames are captured in 8-bit color depths which provides enough a color range to detect abnormalities. Here, superpixel based methods are proposed to segment five diseases including: bleeding, Crohn's disease, Lymphangiectasia, Xanthoma, and Lymphoid hyperplasia. Two superpixels methods are compared to provide semantic segmentation of these prolific diseases: simple linear iterative clustering (SLIC) and quick shift (QS).  The segmented superpixels were classified into two classes (normal and abnormal) by support vector machine (SVM) using texture and color features. For both superpixel methods, the accuracy, specificity, sensitivity, and precision (SLIC, QS) were around 92\%, 93\%, 93\%, and 88\%, respectively. However, SLIC was dramatically faster than QS.
\end{abstract}

\section{Introduction}
\label{sect:intro}
\begin{figure}[t!p]
         \centering
         \begin{subfigure}[b]{0.6\textwidth}
                 \includegraphics[width=1\textwidth]{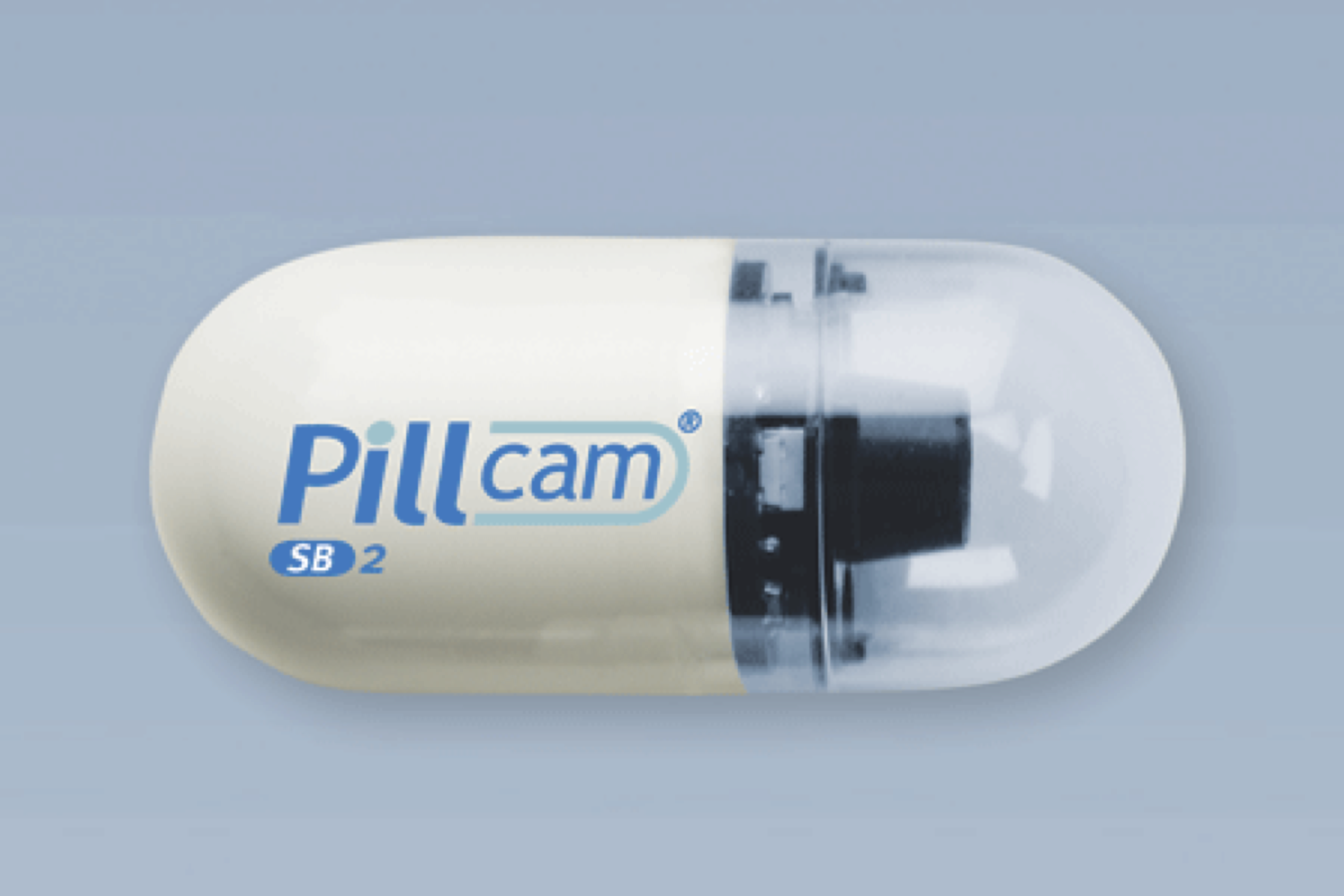}
         \end{subfigure}
\caption{The wireless capsule endoscopy device. Pillcam designed by Given Imaging.} \label{fig1}
\end{figure}

Wireless capsule endoscopy (WCE) is relatively a new device being able to investigate the entire gastrointestinal (GI) tract without any pain. WCE has been used to detect small bowel abnormalities such as a tumor, bleeding, ulcers, Crohn's disease, and celiac. This device is like a normal pill as illustrated in Fig \ref{fig1}. During 8 to 12 hours examination, it records more than 55,000 frames \cite{alizadeh2014segmentation}. To review the recorded frames, a physician needs to spend at least an hour to review all the frames which are a time-consuming process. In addition, this review process can be prone to human error. Image processing has been used frequently to help segmenting, classification, and tracking of objects in different application \cite{li2009computer}, \cite{penjweini2017investigating}, \cite{maghsoudi2017superpixels}; therefore it can address this need.\\
Many Methods have been presented to segment the regions containing abnormalities or to find frames containing those regions. In another word, these methods can be classified into two classes based on the type of detection: regions (pixel) based and frame based. \\
To detect the bleeding in the WCE frames, numerous methods have been presented in recent years. The majority of these methods utilized color and texture features to differentiate the bleeding regions/frames from the normal regions/frames. Combination of following features and classifiers were used to achieve this goal: extracting features from the HSI and the RGB color spaces and classifying by a probabilistic neural network \cite{pan2011bleeding}; extracting chrominance moment and uniform local binary pattern, and classifying by a multilayer perceptron neural network \cite{li2009computer}; extracting Haralick, Gabor, and Law’s texture features and classifying using three neural networks \cite{maghsoudi2016detection}.
Considering 30\% to 50\% of adults might have polyps, they can be considered as one of the most common diseases in the intestine. This change can increase to 90\% for people having more than 50 years old \cite{yuan2016improved}. Most polyps are not cancerous, but if one becomes large enough, there is a chance of turning into cancer. Therefore, it is necessary to detect the polyps in their early stages.\\
In one of the most recent studies, a method was proposed to differentiate the frames containing polyps from the frames having only the normal regions \cite{yuan2016improved}. Single scale-invariant feature transform (SIFT), local binary pattern (LBP), uniform LBP, complete local binary pattern (CLBP), and a histogram of oriented gradients (HOG) were extracted from the WCE frames. Then, a support vector machine (SVM) and Fisher's linear discriminated analysis was used to classify the frames. The highest accuracy, specificity, and sensitivity for detection of frames containing polyp were 93\%, 94\%, and 87\%, respectively.\\
In another study, a pixel-based method was presented to segment the polyps in two steps: first, combined log Gabor filters and Susan edge detector were used to creating potential polyp segments, and then, geometric features were extracted to outline final polyps regions {\cite{karargyris2011detection}}. Edge detection techniques following by Hough transform were used to extract features based on shape. Texture features were added to these features, and finally, a Cascade Adaboost was used to classify the frames using these features. A sensitivity of 91\% and a specificity of 95\% were reported {\cite{silva2014toward}}. \\
There has been some studies to segment or detect other types of diseases in the WCE frames \cite{maghsoudi2016detection}, \cite{maghsoudi2013detection}, \cite{maghsoudi2017superpixel}, \cite{li2011computer}, \cite{kumar2012assessment}, \cite{segui2016generic}, \cite{omid2012segmentation}. A technique was presented to detect Crohn's disease in the WCE frame {\cite{kumar2012assessment}}. The MPEG-7 edge histogram descriptor (edge feature), the MPEG-7 dominant color descriptor clustering colors from the LUV color space, and the MPEG-7 homogeneous texture descriptor using Gabor filters (texture features) were extracted. Finally, an SVM was used to classify the regions using these extracted features. The results indicated an accuracy of 93\%, a precision of 91\%, and a specificity of 91\%. \\
To segment Crohn's disease (Lymphangiectasia, Xanthoma, Lymphoid hyperplasia, and Stenosis), a simple method was presented using a sigmoid function to emphasize on the region based on the intensity value {\cite{omid2012segmentation}}. The sensitivity, specificity, and accuracy were reported by 89\%, 65\%, and 75\%, respectively. \\
In addition, methods were proposed to detect frames containing tumor regions in the WCE frames {\cite{maghsoudi2016detection}}, {\cite{li2012tumor}}. \\
Here, two superpixel based methods, simple linear iterative clustering (SLIC) and quick shift (QS), are presented to segment the bleeding, Crohn's disease, Lymphangiectasia, Xanthoma, and Lymphoid hyperplasia regions in the WCE frames. Then, an SVM is used to classify the segmented regions. The main contribution of this work is to demonstrate how SLIC and QS superpixels can be used to segment the diseases regions in the WCE frames. The accurate segmentation leads to a superior classification of these regions, especially for some of these diseases compared to the previous studies. 
\section{Method}
\label{sect:meth}
\subsection{SLIC Superpixel}
Superpixels contract and group uniform pixels in an image and it has been so popular for different applications like segmentation, object recognition, and tracking of objects in a video \cite{Ren03}, \cite{Comaniciu02}, \cite{Felzenszwalb04}, \cite{Levinshtein09}. The main idea for superpixels were presented as defining the perceptually uniform regions using the normalized cuts algorithm \cite{shi2000normalized}, \cite{Ren03}, \cite{Mori04}, \cite{Li12} . Superpixels create a more natural and perceptually meaningful representation of an image compared to the many other methods available for segmentation. \\ 
Here, we used SLIC {\cite{Achanta12}} and QS {\cite{vedaldi2008quick}} to evaluate the superpixel segmentation performance to segment the five diseases regions in the WCE frames. SLIC can be considered as a form of k-means clustering but it has two main differences: the number of distance calculations is decreased by superpixels size, color, and spatial relations are combined to update the size and compactness of superpixels {\cite{Mori04}}. \\
The key parameter for SLIC is the number of superpixels. First, $N$ centers are considered as the cluster centers. Then, to avoid keeping the center located on the edge (high gradient), the center is transferred to the lowest gradient position in a $3\times3$ neighborhood. Each of the pixels is associated with the nearest cluster center based on color information. Therefore, two coordinate components ($x$ and $y$) show the location of the segment and three color components (for example $R$, $G$, and $B$ intensities in the RGB color space) are derived. SLIC finds and minimizes a distance (a Euclidean norm on 5D spaces) function defined as follow:
\begin{equation}
\label{eq:1}
D_{c} = \sqrt{(R_{j}-R_{i})^{2}+(G_{j}-G_{i})^{2}+(B_{j}-B_{i})^{2}},
\end{equation}
\begin{equation}
\label{eq:2}
D_{p} = \sqrt{(x_{j}-x_{i})^{2}+(y_{j}-y_{i})^{2}},
\end{equation}
\begin{equation}
\label{eq:3}
D = \sqrt{(\frac{D_{c}}{N_{c}})^{2}+(\frac{D_{p}}{N_{p}})^{2}}.
\end{equation}
Where $N_{c}$ and $N_{p}$ are maximum distances within a cluster used to normalize the color and spatial proximity. It should be said that SLIC keeps the size of superpixels between half and twice of the initially specified superpixel size. Therefore, the number of superpixels for the SLIC method determines an important role on how the segmentation can be performed. This effect is illustrated in Fig \ref{fig3}. 
\subsection{Features}
Segmentation and detection of the five common diseases (bleeding, Crohn's disease, Lymphangiectasia, Xanthoma, and Lymphoid hyperplasia) in the GI tract were the main goal of this study. Therefore, we needed to find the best possible features differentiating these five diseases regions from the normal landmarks in the frames. Because the shape of the diseases was varying dramatically from a frame to another frame, the shape and the size of superpixels cannot provide distinctive features. On the other hand, color and texture features can provide enough information for this distinction. \\
Local binary pattern has been widely used to extract texture feature \cite{yuan2016improved}, \cite{maghsoudi2016detection}, \cite{ojala2002multiresolution}, \cite{nawarathna2014abnormal}. The superiority of uniform LBP to the other texture feature methods is that it is invariant to rotation and scaling {\cite{ojala2002multiresolution}}.
\begin{figure*}[t]
         \centering
         \begin{subfigure}[b]{0.8\textwidth}
                 \includegraphics[width=1\textwidth]{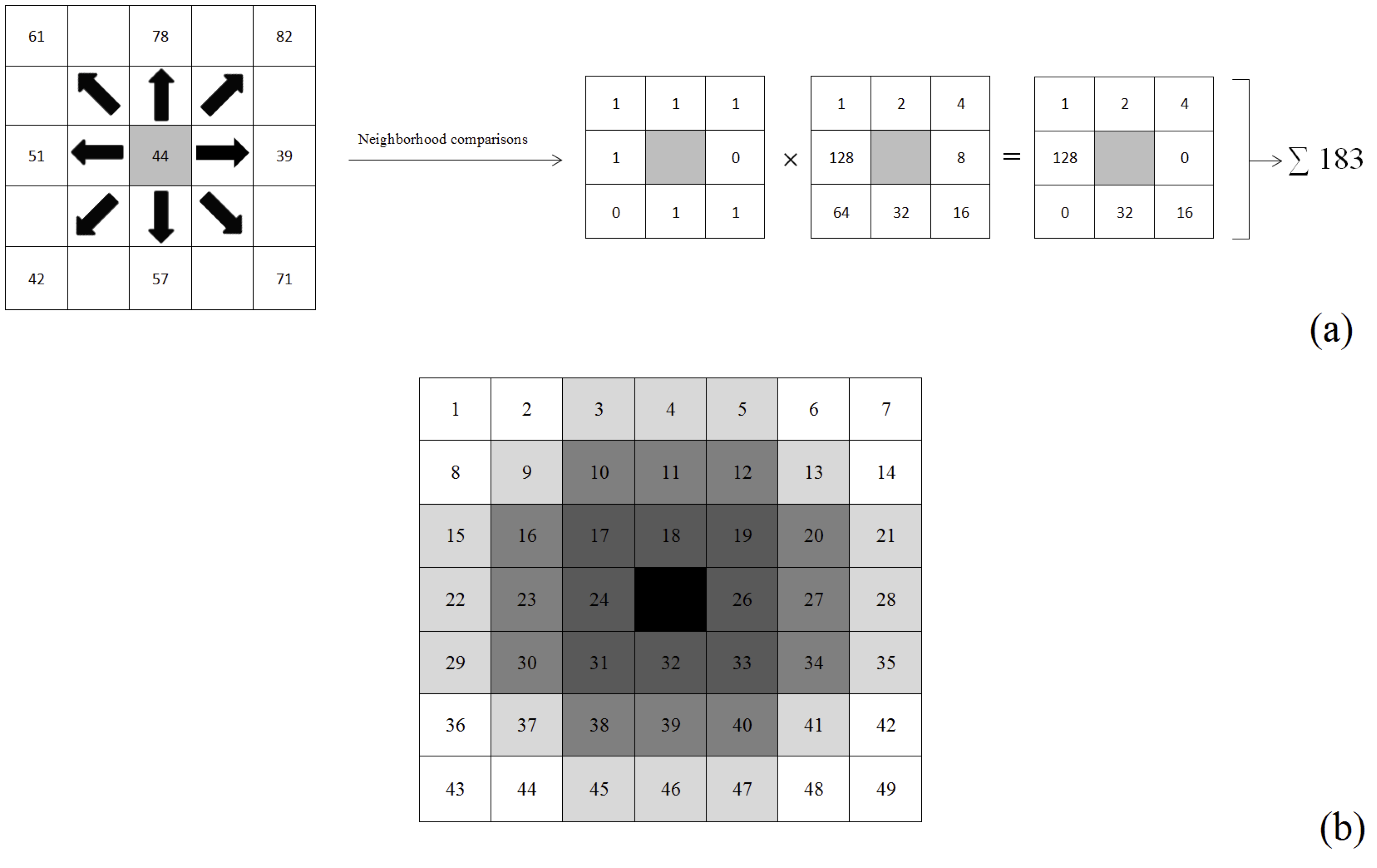}
         \end{subfigure}
\caption{LBP calculation for an example block. (a) shows how the LBP can be calculated from a block. (b) demonstrates how different radii (1, 2, and 3) are chosen for the LBP calculation.} \label{fig2}
\end{figure*}
To calculate the LBP, a function \textit{T} can be defined as follow for \textit{n} neighbors, \\
\begin{equation}
    T=t(g_{c}, g_{0}, ..., g_{n-1})
\end{equation}
where $g_{c}$ shows the center pixel intensity, and $g_{p}$(\textit{p} = 0, . . ., \textit{n}-1) is the intensity value of the pixels locating on a circle with a radius of \textit{R}. The coordinates of these neighbors can be given by ($x_{c}$+\textit{R}cos(2$\pi$\textit{p}/\textit{n}), $y_{c}$-\textit{R}sin(2$\pi$\textit{p}/\textit{n})), in which $x_{c}$, $y_{c}$ are the coordinates of the pixel located in the center of block. If the intensity of the center is subtracted from the intensities of all the neighbors, then the texture function can be written as: \\
\begin{equation}
    T=t(g_{c}, g_{0}-g_{c}, ..., g_{p-1}-g_{c})
\end{equation}
where $g_{c}$ shows the intensity and the function can be redefined as follows: \\
\begin{equation}
    T=t(s(g_{0}-g_{c}), ..., s(g_{p-1}-g_{c}))
\end{equation}
Finally, the LBP feature can be calculated using the following equation: \\
\begin{equation}
    LBP _{P,R(x_{c}, y_{c})}=\sum^{P-1}_{p=0}s(g_{p-1}-g_{c})\times 2^{p}
\end{equation}
where:
\begin{equation}
    S(x)=\left\{
                \begin{array}{   l   l    }
                  $1\ \ \ \ \ \ \ \ x>1$\\
                  $0\ \ \ \ \ \ \ \ x<0$\\
                \end{array}
              \right.
\end{equation}
How LBP is calculated is illustrated in Fig \ref{fig2} for an example block.\\
In addition to LBP and uniform LBP, we extracted following five measures form the gray scale image, LBP, and uniform LBP: mean, variance, skewness, kurtosis, and entropy. \\
To extract color features, the images were transferred to the HSV color space. The same five features were extracted from the hue, red, green, and blue channels. 
\begin{figure*}[tp]
          \centering
         \begin{subfigure}[b]{1\textwidth}
                 \includegraphics[width=1\textwidth]{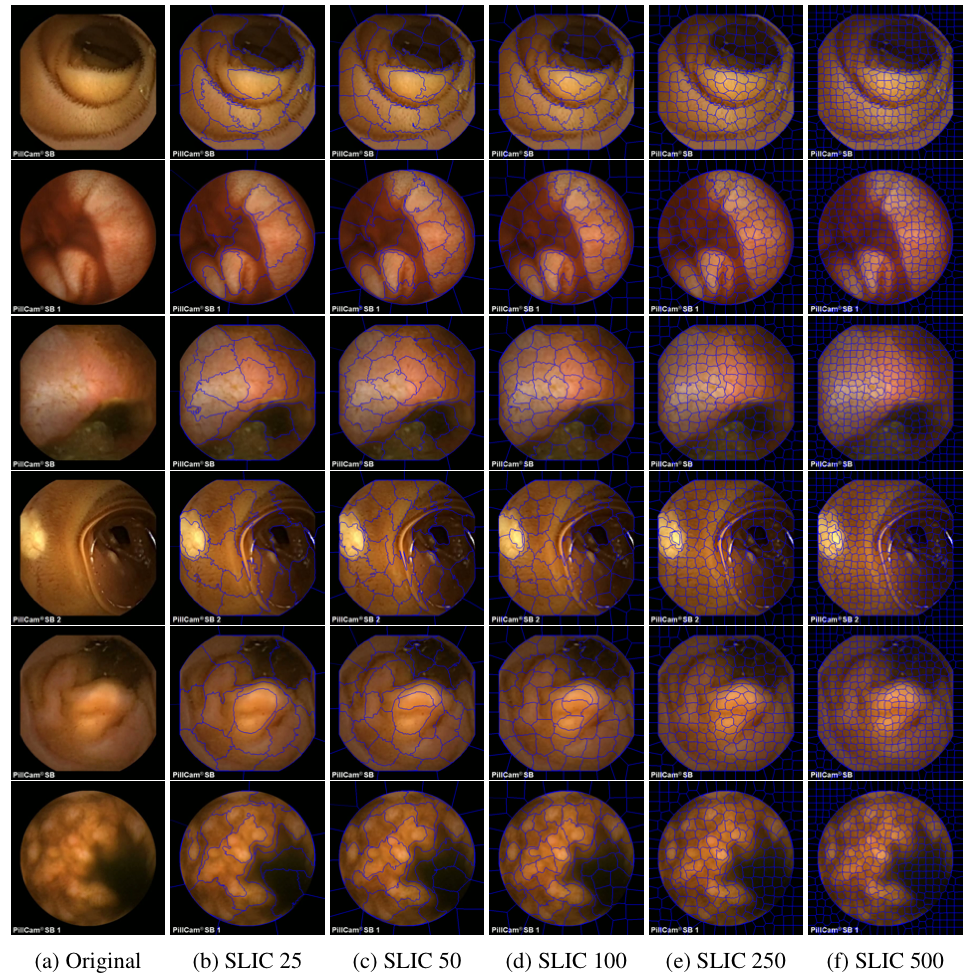}
         \end{subfigure}
\caption{ The number of superpixel effect on SLIC segmentation. A, b, c, d, e, and f columns show the original image, the result for SLIC with a size of 25, 50, 100, 250, and 500. From top to bottom, the rows show normal, bleeding, Crohn's disease, Lymphangiectasia, Xanthoma, and Lymphoid hyperplasia frames.} \label{fig3}
\end{figure*}
SVM has been one of the most popular classifiers in different applications and especially to classify objects in the WCE frames \cite{kumar2012assessment}, \cite{li2012tumor}, \cite{maghsoudi2016tracker}. Therefore, we used SVM to classify the superpixels using the extracted features.
\section{Results}
\label{sect:res}
\begin{figure*}[tp]
          \centering
         \begin{subfigure}[b]{1\textwidth}
                 \includegraphics[width=1\textwidth]{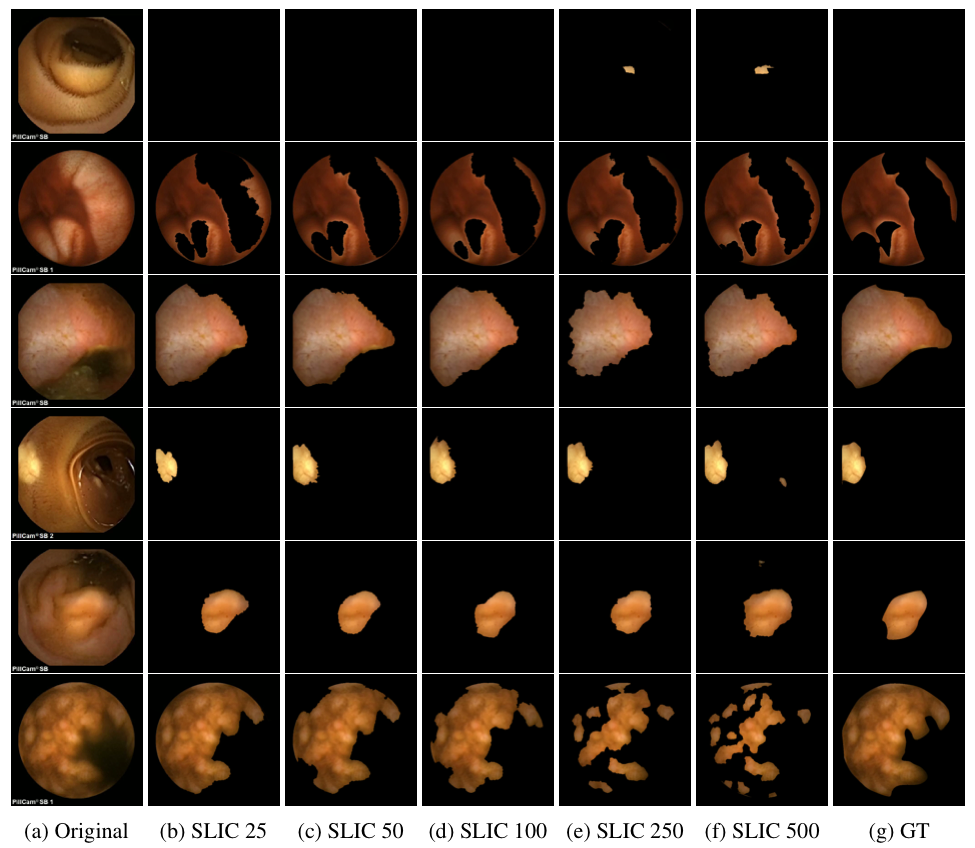}
         \end{subfigure}
\caption{Final segmented and classified regions after using SVM. A, b, c, d, e, and f columns show the original image, the final detected regions for SLIC with 25, 50, 100, 250, and 500 superpixels. The last column, g, shows the grand truth which is manually outlined. From top to bottom, the rows show normal, bleeding, Crohn's disease, Lymphangiectasia, Xanthoma, and Lymphoid hyperplasia frames.} \label{fig4}
\end{figure*}

\begin{figure}[bp]
         \centering
         \begin{subfigure}{0.75\textwidth}
                 \includegraphics[width=1\textwidth]{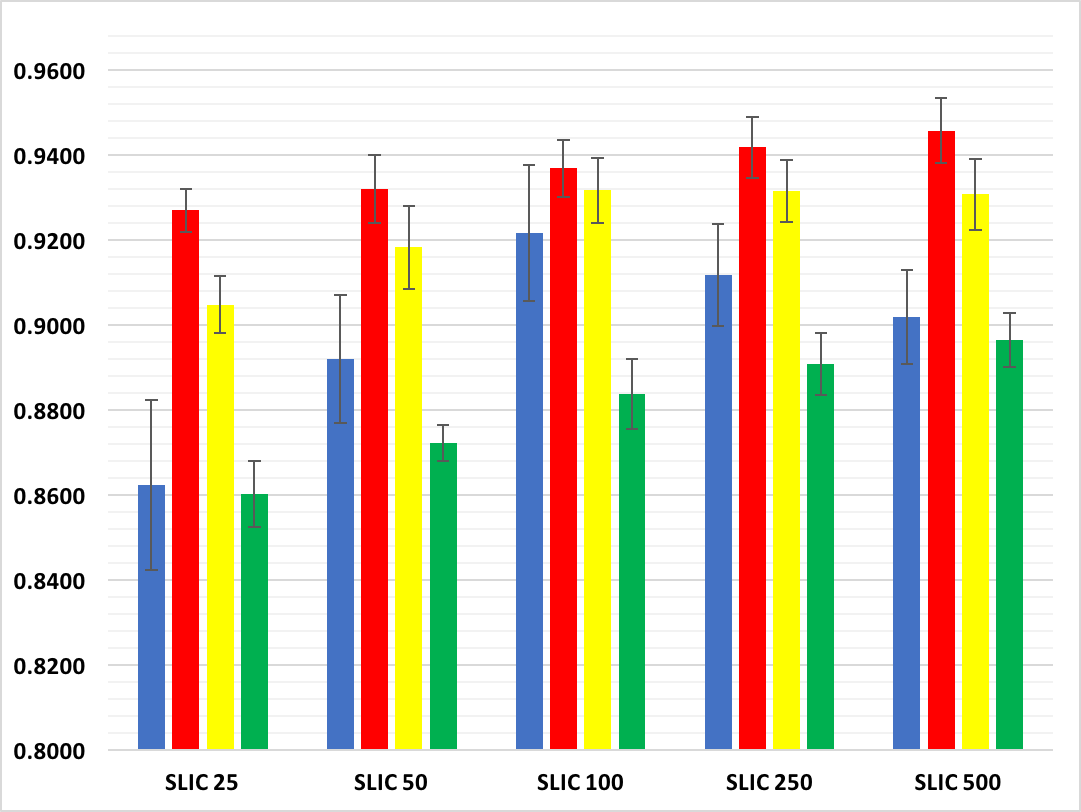}
         \end{subfigure}
         \caption{The bar plot of four measures based on the five SLIC size. The blue, red, black, and green bars are sensitivity, specificity, accuracy, and precision.} \label{fig5}
\end{figure}

To evaluate the proposed methods, 39 frames containing bleeding taken from 9 patients, 28 frames containing Crohn's disease taken from 5 patients, 25 frames containing Lymphangiectasia taken from 4 patients, 19 frames containing Xanthoma taken from 3 patients, and 24 frames containing Lymphoid hyperplasia taken from 4 patients were gathered from the Shariati Hospital, Tehran, Iran. \\
To train the SVM, we randomly selected one of the patients having at least 6 frames from each of diseases. Only for the bleeding class, we had to select two patients because none of the patients had more than 6 frames in our database. The remaining frames were used for testing the trained SVM.\\
Five measures, mean, variance, skewness, kurtosis, and entropy, were extracted from LBP, uniform LBP, gray scale image, hue, red, green, and blue channels. This process created 35 features. We used the Laplacian score test to reduce the number of features and find the best distinctive features {\cite{he2005laplacian}}. \\
As discussed in section \ref{sect:meth}, the main parameter affecting the superpixels was the number of superpixels. The frames collected from the Hospital were in 8-bit color depth and with the resolution of $512\times512$. Five superpixel numbers (25, 50, 100, 250, and 500) were selected to evaluate the size effect for segmentation and classification of diseases. This effect is illustrated in Fig \ref{fig3} and the results are shown in Fig \ref{fig4}. It should be noted that we trained five SVMs for each of the superpixel numbers as the features were different based on the number of superpixels. 
To quantify the segmentation after classification, the accuracy, precision, sensitivity, and specificity were measured as follows: \\
\begin{equation}
Sensitivity=\frac{TP}{TP+FN}
\end{equation}
\begin{equation}
Specificity=\frac{TN}{TN+FP}
\end{equation}
\begin{equation}
Accuracy=\frac{TP+TN}{TP+FN+TN+FP}
\end{equation}
\begin{equation}
Precision=\frac{TP}{TP+FP}
\end{equation}
where TP, FN, TN, and FP denote the number of pixels in abnormal regions that were correctly labeled, the number of pixels in abnormal regions that were incorrectly labeled as normal, the number of pixels in normal tissue regions that were correctly labeled, and the number of pixels in normal tissue regions that were incorrectly labeled as abnormal.\\
\begin{table*}[tp]
  \centering
  \caption{The sensitivity, specificity, accuracy, and precision for all five diseases and separately. The results are illustrated in six parts: total, bleeding, Crohn's disease, Lymphangiectasia, Xanthoma, and Lymphoid hyperplasia. Each of these parts shows the five SLIC number of superpixels (25, 50, 100, 250, and 500).} \label{tab1}
\scalebox{0.45}{
    \begin{tabular}{|c|c|c|c|c|c|c|c|c|c|c|c|}
    \toprule
    \textbf{Measures} & \textbf{SLIC 25} & \textbf{SLIC 50} & \textbf{SLIC 100} & \textbf{SLIC 250} & \textbf{SLIC 500} & \textbf{Measures} & \textbf{SLIC 25} & \textbf{SLIC 50} & \textbf{SLIC 100} & \textbf{SLIC 250} & \textbf{SLIC 500} \\
    \midrule
    \multicolumn{6}{|c|}{\textbf{Total}}          & \multicolumn{6}{c|}{\textbf{Lymphangiectasia}} \\
    \midrule
    \textbf{Sensitivity} & \textbf{0.8623} & \textbf{0.8920} & \textbf{0.9218} & \textbf{0.9119} & \textbf{0.9020} & \textbf{Sensitivity} & \textbf{0.8910} & \textbf{0.8940} & \textbf{0.9029} & \textbf{0.9094} & \textbf{0.9049} \\
    \midrule
    \textbf{Specificity} & \textbf{0.9270} & \textbf{0.9320} & \textbf{0.9369} & \textbf{0.9418} & \textbf{0.9458} & \textbf{Specificity} & \textbf{0.9053} & \textbf{0.9191} & \textbf{0.9162} & \textbf{0.9152} & \textbf{0.9182} \\
    \midrule
    \textbf{Accuracy} & \textbf{0.9049} & \textbf{0.9183} & \textbf{0.9317} & \textbf{0.9316} & \textbf{0.9308} & \textbf{Accuracy} & \textbf{0.9004} & \textbf{0.9105} & \textbf{0.9116} & \textbf{0.9132} & \textbf{0.9136} \\
    \midrule
    \textbf{Precision} & \textbf{0.8602} & \textbf{0.8722} & \textbf{0.8838} & \textbf{0.8909} & \textbf{0.8965} & \textbf{Precision} & \textbf{0.8306} & \textbf{0.8520} & \textbf{0.8487} & \textbf{0.8481} & \textbf{0.8520} \\
    \midrule
    \multicolumn{6}{|c|}{\textbf{Bleeding}}       & \multicolumn{6}{c|}{\textbf{Xanthoma}} \\
    \midrule
    \textbf{Sensitivity} & \textbf{0.9594} & \textbf{0.9664} & \textbf{0.9713} & \textbf{0.9693} & \textbf{0.9644} & \textbf{Sensitivity} & \textbf{0.8920} & \textbf{0.9000} & \textbf{0.9069} & \textbf{0.9119} & \textbf{0.9049} \\
    \midrule
    \textbf{Specificity} & \textbf{0.9645} & \textbf{0.9685} & \textbf{0.9665} & \textbf{0.9665} & \textbf{0.9675} & \textbf{Specificity} & \textbf{0.9182} & \textbf{0.9290} & \textbf{0.9349} & \textbf{0.9379} & \textbf{0.9389} \\
    \midrule
    \textbf{Accuracy} & \textbf{0.9628} & \textbf{0.9677} & \textbf{0.9681} & \textbf{0.9675} & \textbf{0.9664} & \textbf{Accuracy} & \textbf{0.9092} & \textbf{0.9191} & \textbf{0.9253} & \textbf{0.9290} & \textbf{0.9272} \\
    \midrule
    \textbf{Precision} & \textbf{0.9337} & \textbf{0.9410} & \textbf{0.9379} & \textbf{0.9377} & \textbf{0.9392} & \textbf{Precision} & \textbf{0.8502} & \textbf{0.8684} & \textbf{0.8789} & \textbf{0.8843} & \textbf{0.8852} \\
    \midrule
    \multicolumn{6}{|c|}{\textbf{Crohn}}          & \multicolumn{6}{c|}{\textbf{Lymphoid hyperplasia}} \\
    \midrule
    \textbf{Sensitivity} & \textbf{0.9129} & \textbf{0.9049} & \textbf{0.9119} & \textbf{0.9129} & \textbf{0.9158} & \textbf{Sensitivity} & \textbf{0.9168} & \textbf{0.9317} & \textbf{0.9238} & \textbf{0.8940} & \textbf{0.8524} \\
    \midrule
    \textbf{Specificity} & \textbf{0.9142} & \textbf{0.9152} & \textbf{0.9172} & \textbf{0.9142} & \textbf{0.9073} & \textbf{Specificity} & \textbf{0.9369} & \textbf{0.9468} & \textbf{0.9389} & \textbf{0.9517} & \textbf{0.9389} \\
    \midrule
    \textbf{Accuracy} & \textbf{0.9137} & \textbf{0.9117} & \textbf{0.9154} & \textbf{0.9137} & \textbf{0.9102} & \textbf{Accuracy} & \textbf{0.9300} & \textbf{0.9416} & \textbf{0.9337} & \textbf{0.9319} & \textbf{0.9093} \\
    \midrule
    \textbf{Precision} & \textbf{0.8471} & \textbf{0.8475} & \textbf{0.8515} & \textbf{0.8471} & \textbf{0.8373} & \textbf{Precision} & \textbf{0.8833} & \textbf{0.9011} & \textbf{0.8872} & \textbf{0.9060} & \textbf{0.8789} \\
    \bottomrule
    \end{tabular}
}
\end{table*}
The results are quantified in Table \ref{tab1} showing how the SLIC superpixels  method was preformed to segment the regions and SVM labeled the segmented superpixels using the extracted features. Table \ref{tab1} consists of six parts: results reported on top left of table showing the result when all five diseases were considered as a class of abnormal, and the other five parts of the table showing the results for each of the five diseases. The labeled superpixels was compared to a manual outlined region in the frames. The results for labeling of all five diseases are illustrated in Fig \ref{fig5}. \\
Amongst superpixels methods, QS algorithm was selected to be compared with the SLIC results. A 2.7 GHz intel core i5 MacBook pro with 8GB 1867MHz DDR3 memory was used to perform the methods in Python 3. Fig \ref{fig6} compares the four measures (sensitivity, specificity, accuracy, and precision) for segmentation and classification of the superpixels between SLIC and QS algorithms. In addition, to evaluate the speed of the methods, we compared the required average time for segmentation using SLIC method with the QS approach; this comparison is illustrated in Fig \ref{fig7}. 
\section{Discussion}
\label{sect:dis}
The need for segmentation and detection of diseases in the WCE frames has been discussed \cite{schwartz2007small}, \cite{eliakim2013video}. In addition, it helps physicians to review the frames by an accurate finding of the regions in the WCE frames. To address this need, it is vital to segment the regions accurately, and then, to classify the segmented regions. The  SLIC and QS methods presented here can segment five diseases, bleeding, Crohn's disease, Lymphangiectasia, Xanthoma, and Lymphoid hyperplasia, in the WCE frames. Then, the segmented regions were classified using trained SVMs. 
\begin{figure}[b!p]
         \centering
         \begin{subfigure}{0.75\textwidth}
                 \includegraphics[width=1\textwidth]{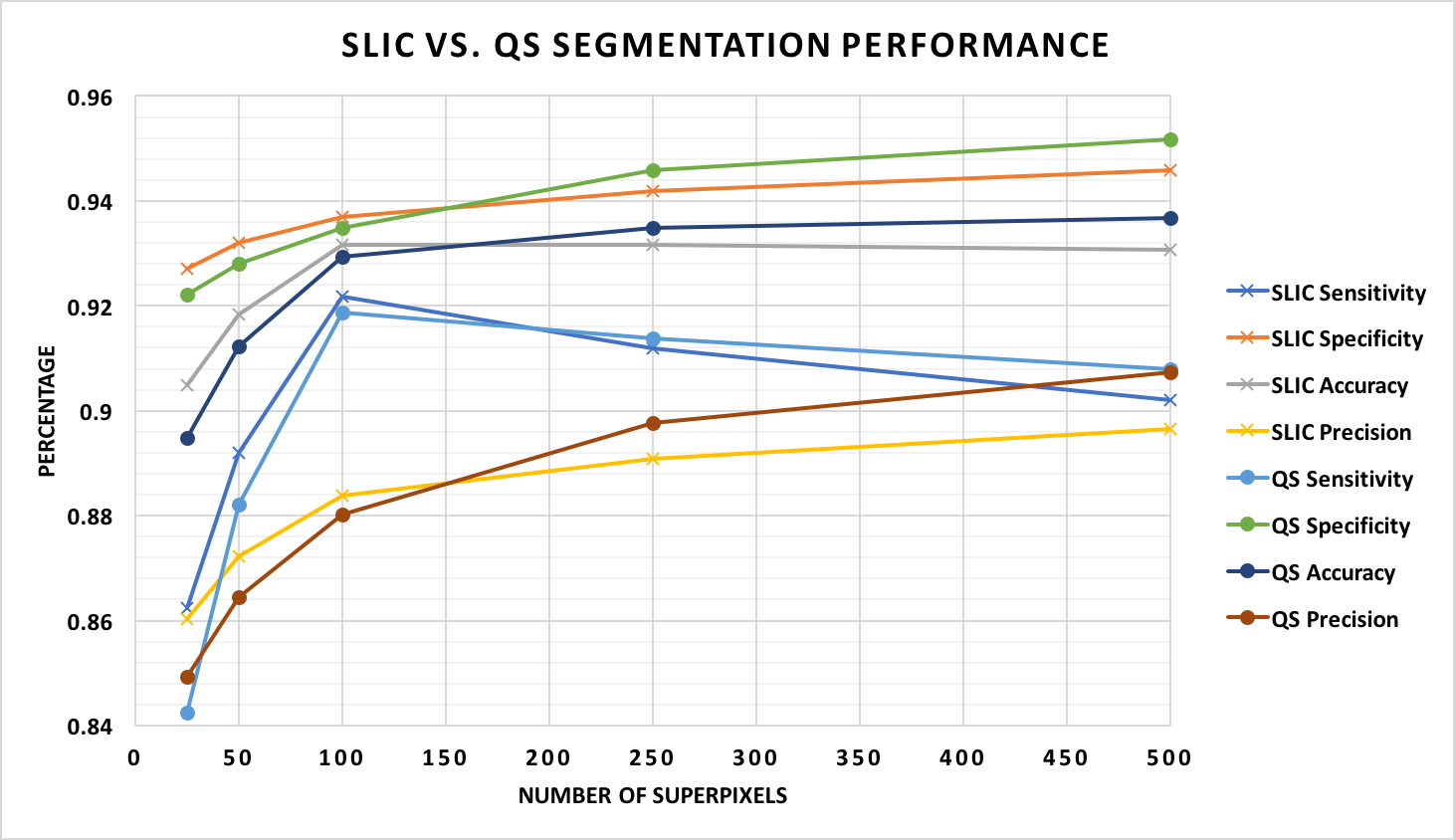}
         \end{subfigure}
         \caption{Comparison of the SLIC and QS methods for segmentation and classification. This graph shows the four measures (sensitivity, specificity, accuracy, and precision) to quantify the performance of SLIC and QS. QS achieved slightly better results when having more superpixels while SLIC was more accurate for fewer superpixels.} \label{fig6}
\end{figure}
Bleeding can be considered as one of a common type of diseases in the GI tract and it can cause a more severe disease. The majority of studies have been devoted to the segmentation and detection of bleeding in the WEC frames \cite{pan2011bleeding}, \cite{li2009computer}, \cite{maghsoudi2016detection}, \cite{karargyris2011detection}, \cite{guobing2011novel}. \\
Most of these methods tried to find the frames containing bleeding regions, frame-based methods, and a few tried to detect the regions in the frames, pixel based methods. The sensitivity of a pixel-based method was reported more than 92\% and more than 98\% for frame-based detection of bleeding {\cite{pan2011bleeding}}. We achieved to a sensitivity of 97\% for detection bleeding regions as reported in Table \ref{tab1}. An improvement of 5 percent will help us to develop the methods for a frame based approach in future works by gathering more frames.\\
To segment the Crohn's disease, a method proposed by Kumar {\cite{kumar2012assessment}} achieved to an accuracy of 93\% and precision of 91\%. Another method tried to segment this type of disease, but the results showed an accuracy of 75\% {\cite{maghsoudi2012segmentation}}. The sensitivity, accuracy, precision, and specificity of our method were separately measured for each of the five diseases. These four measures for detection of Crohn's disease were 91\%, 91\%, 91\%, and 85\%, respectively. Our proposed methods achieved to slightly better results compared to the previous methods for this type of disease. \\
To segment the Lymphangiectasia region in WCE frames, a method was proposed {\cite{cui2010detection}}. The method was dependent on the size of the disease region in the frames. The accuracy and sensitivity were reported 94\% and 55\%, respectively. Our method showed a more accurate detection by having the accuracy and sensitivity of 91\% and 90\% which were significantly better than the previous methods. A summary of available methods compared to our proposed methods is illustrated in Table \ref{tab2}.
\begin{figure}[tp]
         \centering
         \begin{subfigure}{0.75\textwidth}
                 \includegraphics[width=1\textwidth]{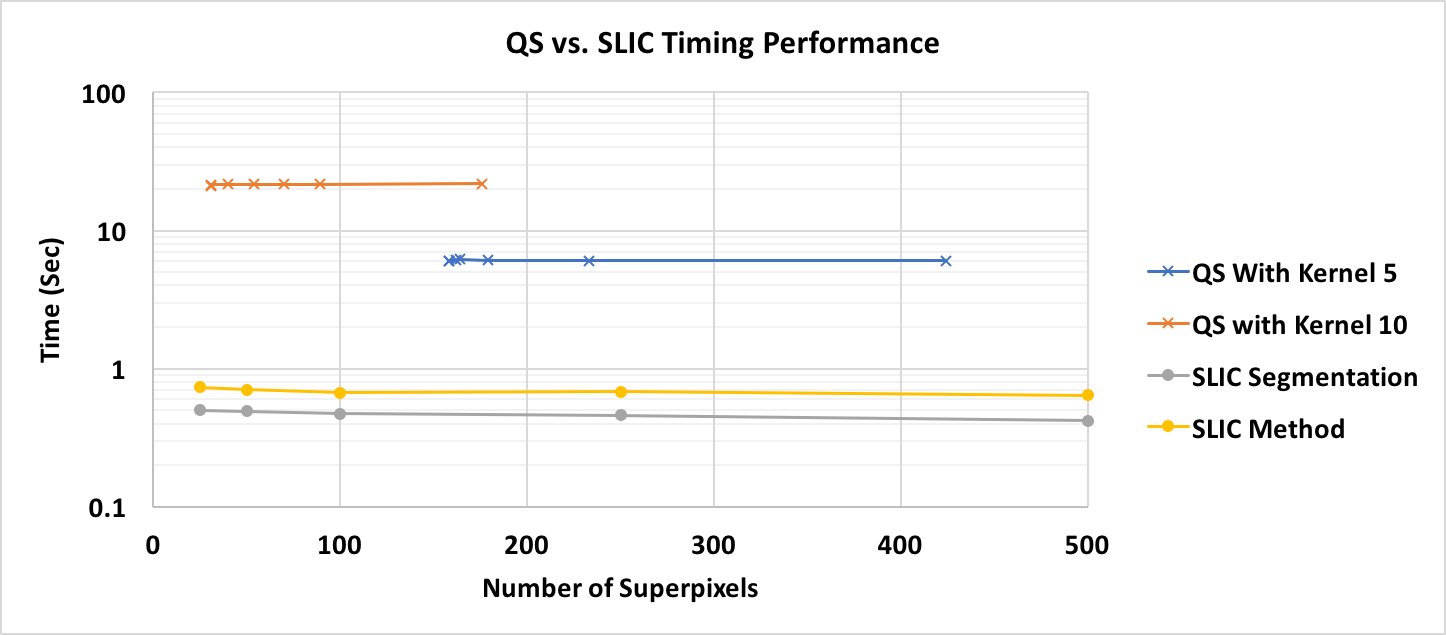}
         \end{subfigure}
         \caption{The required average time for QS and SLIC to segment the superpixels. QS had two parameters to change the number of superpixels; kernel size and maximum distance. The graphs for QS shows the average required time for two kernel sizes (5 and 10). Also, the points on the QS plots from left to right show the QS results for a maximum distance of 10, 15, 20, 25, 30, 100, and 1000. The SLIC segmentation graph shows the average required time for just the segmentation and the SLIC method graph illustrate the average required time for running the whole method.} \label{fig7}
\end{figure}
As Fig \ref{fig3}, Fig \ref{fig4}, and Table \ref{tab1} show the SLIC based method achieved reliable results for detection of the five diseases regions in the WCE frames. The results show the trend of accuracy was a slightly different from one disease to another by changing the number of superpixels. After, reviewing the results and checking the frames, it was discovered that the diseases by a larger area showed a more dramatic decrease in the results after a specific number of superpixels. This trend was seen in Crohn's disease, Lymphoid hyperplasia, and bleeding as they usually have a region with more than 25\% of the whole image pixels. While the other two diseases, Lymphangiectasia and Xanthoma, had smaller regions and the results increased by growing the number of superpixels. 
\begin{table}[tp]
\caption{Comparison between the available methods and the proposed method using SLIC. The first column shows the method type which can be classification after segmentation or only segmentation. FB, PB, and NR are abbreviations for frame-based method, pixel-based, and results which have not reported.} 
\label{tab2}
\begin{center}   
\scalebox{0.8}{
\begin{tabular}{| c | c | c | c | c | c | c | c |}
    \hline
    Method & Mode & Aim & Sensitivity & Specificity & Accuracy & Precision\\ \hline
       Classification{\cite{pan2011bleeding}} & FB & Bleeding & 0.97 & 0.91 & NR & NR \\ \hline
    Classification{\cite{pan2011bleeding}} & PB & Bleeding & 0.92 & 0.88 & NR & NR \\ \hline
    Classification{\cite{maghsoudi2016detection}} & FB & Polyps & 0.87 & 0.94 & 0.93 & NR \\ \hline
       Classification{\cite{karargyris2011detection}} & FB & Polyps & 1.00 & 0.68 & NR & NR \\ \hline
       Classification{\cite{karargyris2011detection}} & FB & Ulcer & 0.75 & 0.73 & NR & NR \\ \hline
       Classification{\cite{karargyris2011detection}} & PB & Ulcer & 0.88 & 0.84 & NR & NR \\ \hline
       Classification{\cite{karargyris2011detection}} & PB & Polyps & 0.96 & 0.70 & NR & NR \\ \hline
       Classification{\cite{silva2014toward}} & PB & Polyps & 0.91 & 0.95 & NR & NR \\ \hline
       Classification {\cite{li2011computer}} & FB & Tumor & 0.85 & 0.80 & 0.82 & NR \\ \hline
    Classification{\cite{li2011computer}} & FB & Tumor & 0.87 & 0.84 & 0.86 & NR \\ \hline
    Classification{\cite{li2011computer}} & FB & Tumor & 0.88 & NR & 0.86 & 0.88 \\ \hline
       Classification{\cite{kumar2012assessment}} & PB & Crohn's & NR & 0.91 & 0.93 & 0.91 \\ \hline
    Segmentation{\cite{maghsoudi2012segmentation}} & PB & Crohn's & 0.89 & 0.65 & 0.75 & NR \\ \hline
    Segmentation{\cite{maghsoudi2012segmentation}} & PB & Lymphoid & 0.87 & 0.80 & 0.81 & NR \\ \hline
    Classification{\cite{cui2010detection}} & FB & Lymphoid & 0.94 & NR & 0.55 & NR \\ \hline
    Classification{\cite{yuan2017deep}} & FB & Polyp & 0.98 & NR & NR & NR \\ \hline
    Classification SLIC & PB & Bleeding & 0.97 & 0.97 & 0.97 & 0.94  \\ \hline
    Classification SLIC & PB & Crohn's & 0.92 & 0.91 & 0.91 & 0.84 \\ \hline
    Classification SLIC & PB & Lymphan & 0.91 & 0.91 & 0.91 & 0.85 \\ \hline
    Classification SLIC & PB & Xanthoma & 0.91 & 0.93 & 0.92 & 0.88 \\ \hline
    Classification SLIC & PB & Lymphoid & 0.93 & 0.95 & 0.94 & 0.90 \\ \hline
\end{tabular}
}
\end{center}
\end{table}
In addition to this trend, the precision was slightly higher. The reason can be the fact that larger superpixels carries more texture and color information for the detection and classification of regions. In another word, when the number of superpixels got higher and the size of them got smaller (getting closer to be a pixel again), the detection error was increased after passing a specific number of superpixel (this number was 100).\\
Fig \ref{fig6} shows the comparison between two superpixels algorithms: QS and SLIC. As seen, the QS was slightly better when the number of superpixels was fewer than the SLIC method. Fig \ref{fig7} illustrates the required average time for these two methods to segment the frames indicating that the SLIC method needed an average time of 0.7 second to process the frames (all the steps including segmentation and classification). \\
For future studies, we will try to develop a method to detect frames, a frame based method, by gathering more frames from each of diseases. The superpixels, specially SLIC, can provide a wealth of information to segment the abnormal regions in the WCE frames. Other methods like deep learning can be used to improve the results. {\cite{yuan2017deep}}.
\bibliographystyle{spmpsci.bst} 
\bibliography{report}

\begin{thebibliography}{10}
\providecommand{\url}[1]{{#1}}
\providecommand{\urlprefix}{URL }
\expandafter\ifx\csname urlstyle\endcsname\relax
  \providecommand{\doi}[1]{DOI~\discretionary{}{}{}#1}\else
  \providecommand{\doi}{DOI~\discretionary{}{}{}\begingroup
  \urlstyle{rm}\Url}\fi

\bibitem{Achanta12}
Achanta, R., Shaji, A., Lucchi, A., Fua, P., Susstrunk, S.: Slic superpixels
  compared to state-of-the-art superpixel methods.
\newblock IEEE Trans. Pattern Anal. Mach. Intell. \textbf{34}, 2274--2281
  (2012)

\bibitem{alizadeh2014segmentation}
Alizadeh, M., Zadeh, H.S., Maghsoudi, O.H.: Segmentation of small bowel tumors
  in wireless capsule endoscopy using level set method.
\newblock Computer-Based Medical Systems (CBMS), 2014 IEEE 27th International
  Symposium on pp. 562--563 (2014)

\bibitem{Comaniciu02}
Comaniciu, D., Meer, P.: Mean shift: A robust approach toward feature space
  analysis.
\newblock IEEE Trans. Pattern Anal. Mach. Intell. \textbf{24}, 603–619 (2002)

\bibitem{cui2010detection}
Cui, L., Hu, C., Zou, Y., Song, S., He, Q., Meng, M.Q.H.: Detection of
  lymphangiectasia disease from wireless capsule endoscopy images with adaptive
  threshold.
\newblock In: Intelligent Control and Automation (WCICA), 2010 8th World
  Congress on, pp. 3088--3093. IEEE (2010)

\bibitem{eliakim2013video}
Eliakim, R.: Video capsule endoscopy of the small bowel.
\newblock Current opinion in gastroenterology \textbf{29}(2), 133--139 (2013)

\bibitem{Felzenszwalb04}
Felzenszwalb, P., Huttenlocher, D.: Efficient graph-based image segmentation.
\newblock Int. J. Comput. Vis. \textbf{59}, 167–181 (2004)

\bibitem{guobing2011novel}
Guobing, P., Fang, X., Jiaoliao, C.: A novel algorithm for color similarity
  measurement and the application for bleeding detection in wce.
\newblock International Journal of Image, Graphics and Signal Processing
  \textbf{3}(5), 1 (2011)

\bibitem{he2005laplacian}
He, X., Cai, D., Niyogi, P.: Laplacian score for feature selection.
\newblock In: NIPS, vol. 186, p. 189 (2005)

\bibitem{karargyris2011detection}
Karargyris, A., Bourbakis, N.: Detection of small bowel polyps and ulcers in
  wireless capsule endoscopy videos.
\newblock IEEE transactions on BioMedical Engineering \textbf{58}(10),
  2777--2786 (2011)

\bibitem{kumar2012assessment}
Kumar, R., Zhao, Q., Seshamani, S., Mullin, G., Hager, G., Dassopoulos, T.:
  Assessment of crohn disease lesions in wireless capsule endoscopy images.
\newblock IEEE Transactions on biomedical engineering \textbf{59}(2), 355--362
  (2012)

\bibitem{Levinshtein09}
Levinshtein, A., Stere, A., Kutulakos, K., Fleet, D., Dickinson, S., Siddiqi,
  K.: Turbopixels: Fast superpixels using geometric flows.
\newblock IEEE Trans. Pattern Anal. Mach. Intell. \textbf{31}, 2290–2297
  (2009)

\bibitem{li2009computer}
Li, B., Meng, M.Q.H.: Computer-aided detection of bleeding regions for capsule
  endoscopy images.
\newblock IEEE Transactions on Biomedical Engineering \textbf{56}(4),
  1032--1039 (2009)

\bibitem{li2012tumor}
Li, B., Meng, M.Q.H.: Tumor recognition in wireless capsule endoscopy images
  using textural features and svm-based feature selection.
\newblock IEEE Transactions on Information Technology in Biomedicine
  \textbf{16}(3), 323--329 (2012)

\bibitem{li2011computer}
Li, B., Meng, M.Q.H., Lau, J.Y.: Computer-aided small bowel tumor detection for
  capsule endoscopy.
\newblock Artificial Intelligence in Medicine \textbf{52}(1), 11--16 (2011)

\bibitem{Li12}
Li, Z., Wu, X.M., Chang, S.F.: Segmentation using superpixels: A bipartite
  graph partitioning approach.
\newblock in Proc. IEEE CVPR p. 789–796 (2012)

\bibitem{maghsoudi2016detection}
Maghsoudi, O.H., Alizadeh, M., Mirmomen, M.: A computer aided method to detect
  bleeding, tumor, and diseases regions in wireless capsule endoscopy.
\newblock Signal Processing in Medicine and Biology Symposium (SPMB) pp. 1--6
  (2016)

\bibitem{maghsoudi2013detection}
Maghsoudi, O.H., Soltanian-Zadeh, H.: Detection of abnormalities in wireless
  capsule endoscopy frames using local fuzzy patterns.
\newblock In: Biomedical Engineering (ICBME), 2013 20th Iranian Conference on,
  pp. 286--291. IEEE (2013)

\bibitem{maghsoudi2016tracker}
Maghsoudi, O.H., Tabrizi, A.V., Robertson, B., Shamble, P., Spence", A.J.: A
  rodent paw tracker using support vector machine.
\newblock Signal Processing in Medicine and Biology Symposium (SPMB) pp. 1--2
  (2016)

\bibitem{nawarathna2014abnormal}
Nawarathna, R., Oh, J., Muthukudage, J., Tavanapong, W., Wong, J., De~Groen,
  P.C., Tang, S.J.: Abnormal image detection in endoscopy videos using a filter
  bank and local binary patterns.
\newblock Neurocomputing \textbf{144}, 70--91 (2014)

\bibitem{ojala2002multiresolution}
Ojala, T., Pietikainen, M., Maenpaa, T.: Multiresolution gray-scale and
  rotation invariant texture classification with local binary patterns.
\newblock IEEE Transactions on pattern analysis and machine intelligence
  \textbf{24}(7), 971--987 (2002)

\bibitem{maghsoudi2012segmentation}
Omid, H., Talebpour, A., Soltanian-Zadeh, H., Haji-Maghsoodi, N.: Segmentation
  of crohn, lymphangiectasia, xanthoma, lymphoid hyperplasia and stenosis
  diseases in wce.
\newblock Quality of Life Through Quality of Information: Proceedings of
  MIE2012 \textbf{180}, 143 (2012)

\bibitem{pan2011bleeding}
Pan, G., Yan, G., Qiu, X., Cui, J.: Bleeding detection in wireless capsule
  endoscopy based on probabilistic neural network.
\newblock Journal of medical systems \textbf{35}(6), 1477--1484 (2011)

\bibitem{JPHP:JPHP12779}
Penjweini, R., Deville, S., Haji~Maghsoudi, O., Notelaers, K., Ethirajan, A.,
  Ameloot, M.: Investigating the effect of poly-l-lactic acid nanoparticles
  carrying hypericin on the flow-biased diffusive motion of hela cell
  organelles.
\newblock Journal of Pharmacy and Pharmacology pp. n/a--n/a.
\newblock \doi{10.1111/jphp.12779}.
\newblock \urlprefix\url{http://dx.doi.org/10.1111/jphp.12779}

\bibitem{Mori04}
Ren, G.M.X., Efros, A.A., Malik., J.: Recovering human body configurations:
  Combining segmentation and recognition.
\newblock in Proc. IEEE CVPR p. 326–333 (2004)

\bibitem{Ren03}
Ren, X., Malike, J.: Learning a classification model for segmentation.
\newblock in Proc. 9th IEEE ICCV p. 10–17 (2003)

\bibitem{schwartz2007small}
Schwartz, G.D., Barkin, J.S.: Small-bowel tumors detected by wireless capsule
  endoscopy.
\newblock Digestive diseases and sciences \textbf{52}(4), 1026--1030 (2007)

\bibitem{segui2016generic}
Segu{\'\i}, S., Drozdzal, M., Pascual, G., Radeva, P., Malagelada, C., Azpiroz,
  F., Vitri{\`a}, J.: Generic feature learning for wireless capsule endoscopy
  analysis.
\newblock Computers in Biology and Medicine \textbf{79}, 163--172 (2016)

\bibitem{shi2000normalized}
Shi, J., Malik, J.: Normalized cuts and image segmentation.
\newblock IEEE Transactions on pattern analysis and machine intelligence
  \textbf{22}(8), 888--905 (2000)

\bibitem{silva2014toward}
Silva, J., Histace, A., Romain, O., Dray, X., Granado, B.: Toward embedded
  detection of polyps in wce images for early diagnosis of colorectal cancer.
\newblock International Journal of Computer Assisted Radiology and Surgery
  \textbf{9}(2), 283--293 (2014)

\bibitem{vedaldi2008quick}
Vedaldi, A., Soatto, S.: Quick shift and kernel methods for mode seeking.
\newblock Computer vision--ECCV 2008 pp. 705--718 (2008)

\bibitem{yuan2016improved}
Yuan, Y., Li, B., Meng, M.Q.H.: Improved bag of feature for automatic polyp
  detection in wireless capsule endoscopy images.
\newblock IEEE Transactions on Automation Science and Engineering
  \textbf{13}(2), 529--535 (2016)

\bibitem{yuan2017deep}
Yuan, Y., Meng, M.Q.H.: Deep learning for polyp recognition in wireless capsule
  endoscopy images.
\newblock Medical Physics  (2017)

\end{thebibliography}

\end{document}